# Artificial Cardiac Conduction System: Simulating Heart Function for Advanced Computational Problem Solving

Rebaz Mohammed Dler Omer[1], Nawzad K. Al-Salihi[1], Tarik A. Rashid[1*], Aso M. Aladdin[2], Mokhtar Mohammadi[3], Jafar Majidpour[4]

[1]Computer Science and Engineering Department, University of Kurdistan Hewler, Erbil 44001, Kurdistan Region, Iraq. rebaz.mohammed@ukh.edu.krd; n.al-salihi@ukh.edu.krd; tarik.ahmed@ukh.edu.krd
[2]Computer Science Department, College of Science, Charmo University, 46023 Chamchamal/Sulaimani, Kurdistan region, Iraq. aso.aladdin@chu.edu.iq
[3]Department of Information Technology, Lebanese French University, Erbil 44001, Kurdistan Region, Iraq. mokhtar.mohammadi1@gmail.com
[4]Computer Science Department, University of Raparin, Rania 46012, Iraq. jafar.majidpoor@uor.edu.krd

*Corresponding authors email address: tarik.ahmed@ukh.edu.krd

**Abstract**

This work proposes a novel bio-inspired metaheuristic called Artificial Cardiac Conduction System (ACCS) inspired by the human cardiac conduction system. The ACCS algorithm imitates the functional behaviour of the human heart that generates and sends signals to the heart muscle, initiating it to contract. Four nodes in the myocardium layer participate in generating and controlling heart rate, such as the sinoatrial, atrioventricular, bundle of His, and Purkinje fibres. The mechanism of controlling the heart rate through these four nodes is implemented. The algorithm is then benchmarked on 19 well-known mathematical test functions as it can determine the exploitation and exploration capability of the algorithm. The results are verified by a comparative study with Whale Optimization Algorithm (WOA), Particle Swarm Optimization (PSO), Gravitational Search Algorithm (GSA), Differential Evolution (DE), and Fast Evolutionary Programming (FEP). The algorithm undergoes a rigorous evaluation using the CEC-C06 2019 Benchmark Test Functions, illuminating its adeptness in both exploitation and exploration. Validation ensues through a meticulous comparative analysis involving the Dragonfly Algorithm (DA), WOA, PSO, Lagrange Elementary Optimization (Leo), and the Ant Nesting Algorithm (ANA). The results show that the ACCS algorithm can provide very competitive results compared to these well-known metaheuristics and other conventional methods.

## 1. Introduction

Over the past three decays, metaheuristic optimisation techniques have been drawing many researchers' attention. A metaheuristic is a high-level framework for problem-independent algorithms combining intelligently different approaches to developing heuristic optimization algorithms [1]. Some of these algorithms, such as Artificial Immune System (AIS) [2], Ant Colony Optimization (ACO) [3], Particle Swarm Optimization (PSO) [4] and Dragonfly Algorithm (DA) [5], are surprisingly well-known among scientists from a wide variety of disciplines, not only computer scientists. In addition to a vast number of theoretical works, these optimisation techniques have been utilised in numerous academic disciplines. There is a question here regarding why metaheuristics have become so prevalent. This question has a four-part answer: simplicity, flexibility, derivation-free mechanism, and avoidance of local optima [7].

First, metaheuristics are quite simple due to the originating metaheuristic frameworks from natural phenomena. They have been influenced primarily by straightforward concepts. Typically, the sources of inspiration are physical facts, animal behaviour, or evolutionary concepts. Second, metaheuristics do not have any restrictions when formulating an optimization problem, such as predefining constraints or objective functions of the decision variables [7]. Third, the majority of metaheuristics contain processes that do not require derivation. Metaheuristics optimise problems stochastically, as opposed to gradient-based techniques. The optimisation procedure begins with random solutions, and it is not necessary to calculate the derivative of search spaces to locate the optimal solution. This makes metaheuristics particularly applicable to real-world issues involving costly or uncertain derivative data. Finally, metaheuristics are superior to conventional optimisation techniques in avoiding local optima. Due to the stochastic character of metaheuristics, they can avoid stagnation in local solutions and exhaustively scan the whole search space. The search space of real-world problems is typically unknown and highly complicated, containing many local optima; hence, metaheuristics are excellent possibilities for optimising these problematic real-world problems.

It is worth mentioning that no universal algorithm can guarantee to get the optimal solution for all problems, and this statement has been mathematically proven by the No Free Lunch theorem (NFL) [8]. NFL also proves that there is still room for improvement and enhancement since most algorithms may have good results on some problems while performing poorly on other problems. This motivates our efforts to formulate a novel metaheuristic based on cardiac conduction system behaviour. Generally, metaheuristics can be divided into three fundamental classes based on the method by which solutions are manipulated [9]. Local search metaheuristics search for suitable solutions iteratively by conducting small changes to a single solution. Constructive metaheuristics construct the best possible solutions at each iteration rather than improving complete solutions. Population-based metaheuristics find optimum solutions by combining selecting and existing solutions into a new one called the population. Population-based metaheuristics offer several benefits over single-solution-based algorithms; multiple candidate solutions share information about the search space, resulting in abrupt leaps into the most promising part of the search space. Multiple candidate solutions collaborate to eliminate locally optimum solutions. In general, population-based metaheuristics are more exploratory than single-solution-based algorithms.

One of the main branches of population-based metaheuristics is bio-inspired algorithms. Bio-inspired algorithms are inspired by scientific methods that conduct biological metaphors to find and develop new engineering solutions to many complex optimisation and real-world problems. Some of the most popular bio-inspired algorithms are the Whale Optimisation Algorithm (WOA) [10], PSO [4], AIS [2], Genetic Algorithm (GA) [11], Artificial Bee Colony (ABC) [12], Lagrange Elementary Optimization (Leo) [13], and Ant Nesting Algorithm (ANA) [14]. A thorough literature review of bio-inspired algorithms is presented in the next section. Regardless of the differences between the metaheuristics, exploration, and exploitation are the most important two key factors common in all bio-inspired metaheuristics algorithms. Exploration randomly searches for a new solution space to find global optima and exploitation leads to finding local optima in the explored solution space [15]. Hence, the balance between these two vital processes in all bio-inspired algorithms is vital because intense exploration and deep exploitation may have an effect on effect not giving optimal solutions and traps in local optima of algorithms, respectively. This work proposes a new bio-inspired algorithm inspired by the human cardiac conduction system.

The rest of the paper is organised as follows: Section 2 presents a literature review of bio-inspired algorithms. Section 3 states the biological motivation of ACCS. Section 4 outlines the proposed ACCS algorithm. The experimental performance evaluation and discussion of benchmark functions are presented in Section 5. Finally, Section 6 concludes the paper and makes recommendations for future studies.

## 2. Literature Review

Metaheuristics nature-inspired optimisation algorithms are classified into three categories: physics-based algorithms, Swarm Intelligence (SI) algorithms, and bio-inspired (non-SI-based) algorithms. Physical-based optimisation algorithms typically imitate physical rules. Some of the most popular algorithms are Simulated Annealing (SA) [16], Harmony Search (HS) [17], Gravitational Search Algorithm (GSA) [18], Big-Bang Big-Crunch (BBBC) [19], Black Hole (BH) [20] algorithm, and Curved Space Optimization (CSO) [21]. These algorithms receive their inspiration from physical laws such as using ray casting, inertia forces, weights, electromagnetic force, and so on.

For example, the SA algorithm uses the concept of annealing within the metallurgy. It is referred to as the process used to modify certain alloys of metal by heating more than their melting points, holding their temperature, and then regulating the cooling process to enhance its ductility and hardness. The procedural idea behind the SA algorithm is that in the same way the metal atom attempts to reach its ideal state during the cooling down after the annealing process, SA will provide an ideal (best) solution. Similarly, the annealing process discards all solutions that are not ideal for the metal; instead, the SA algorithm uses the probability function to decide whether to discard the inferior solution or accept it. With the metal cooling down the process, the atoms are more likely to reach their ideal state than the state where the metal is in the melting point state. Thus, it can be described SA algorithm in the terminology of metallurgy that, it starts in the hot state and cooling down over time (convergence state) [22]. GSA is another physics-based algorithm. Rashedi, Nezamabadi-pour, and Saryazdinin proposed it in 2009. It is inspired by the Newtonian law of gravity and mass interactions. The GSA method conducts a search using a collection of agents whose masses are proportional to the value of a fitness function. Throughout iteration, the masses are drawn to one another by gravitational forces. The attracting force increases proportionally with mass. Therefore, the heaviest mass, which may be near to the global optimum, attracts the other masses proportionally to their distances.

The second main branch of metaheuristics is SI algorithms. These algorithms mostly imitate the social behavior of the insects like bees, ants, and termites, as well as the social behavior of other animals such as a group of fish or birds. The technique is nearly identical to a physics-based algorithm, but the search agents navigate by simulating organisms' collective and social intelligence. Some of the most popular SI algorithms are PSO [4], ACO [3], DA [13], WOA [9], ABC [11], ANA [14] and Firefly Algorithm (FA) [23]. The PSO algorithm was proposed by Kennedy and Eberhart in 1995 and was motivated by the social behaviour of the birds flying as a flock searching for food. The population of birds moves on some specific path to arrive at their food destination, and the shortest path defined by a bird is considered the best local known solution position or particle best solution. In other words, the movement of a particle takes into account both its optimal solution and the optimal solution obtained by the swarm.

Another popular SI algorithm is ACO, first introduced by Dorigo et al., [3]. The basic concept of this algorithm came from ants, small creatures that can search intelligently for a common source of food through a simple indirect low-level communication channel called pheromone trails. The possible solutions evolve a pheromone matrix by repeated iteration. This characteristic enables ants to be seen as multiple agent procedures that solve complex optimisation problems. The WOA is another popular algorithm rooted in the hunting behaviour of humpback whales. Mirjalili and Lewis developed it in 2016 [10]. The observed and considered behaviors in humpback whales are the bubble-net and spiral bubble-net feeding which have been mathematically modeled to obtain the WOA algorithm. The ABC is another popular SI algorithm proposed by Karaboga in 2005 [12]. Foraging patterns of honeybees inspire the algorithm during the collection of nectar from a flower as a feeding source for the hive. ABC algorithm uses three groups of bees to analyse the algorithm's working procedure: onlookers, scouts, and employed bees. The employed and onlooker bees perform the exploitation process in the search space, while scouts control only the exploration process in the search space. Thus, the scouts search for the source of food randomly, and, and they transfer the nectar amount of each specific food position to the onlookers.

The third subclass of metaheuristics nature-inspired algorithms is the bio-inspired (not SI-based algorithm). These algorithms mostly mimic the biological metaphors to find and develop new engineering solutions to many complex optimisation and real-world problems since it demonstrates robust heterogeneous and fascinating phenomena. The most popular algorithm in this branch is GA. This algorithm was introduced by Holland in 1992 [11] and acquired its motivation from the process of evolution and procreation in biology and the theory of evolution presented by Charles Darwin. The idea behind this algorithm is the possibility of combining two solutions called parents into new solutions called children. The children compared each other with how they will get some features from their parents, and the repetition of this comparison occurs until the best children (i.e. optimal solution) can be indicated as the better solution each child inherited from the best parts of their parents to improve the one started with, and this process called optimisation.

Another popular bio-inspired algorithm is the Bacterial Foraging Optimization Algorithm (BFOA), initially presented by Kevin M. Passino in 2002 [24]. During foraging, biologists discovered that the movement of the bacteria is not random, but it moves toward certain stimuli and far away from others, which is known as bacterial chemotaxis. This behaviour becomes the base for developing the BFOA. The Differential Evolution (DE) is another algorithm initially developed by Storn and Price in 1995 [25]. The DE concept is similar to GA because both use individual populations to look for the optimal solution for an optimisation problem. On the other hand, the leading principal difference between GA, Leo, and DE Algorithm is that GA depends on the crossover, a probability method to exchange information via solutions to locate a better solution, while DE strategies use mutation as the primary search method to find a better solution. Besides, the main operators of Leo and DE are mutation and selection. At the start of the evolution process, the exploration occurs through a mutation operator, and then the mutation operator enables exploitation. Thus, the mutation will be incremented till the best optimal value is satisfied based on the stage of the evolutionary process.

To sum up, despite very intense work accomplished on developing new algorithms, metaheuristic algorithms are still considered to be recently compared to well-established conventional algorithms.

Besides the advantages and peak performance of the current mentioned nature-inspired algorithms, it can be said that each algorithm has its drawback that motivates researchers to investigate and look forward to developing new nature-inspired metaheuristics algorithms. To the best of our knowledge, however, there is no bio-inspired algorithm in the literature mimicking the functional behaviour of the human cardiac conduction system. This motivated our attempt to mathematically model the involuntary behaviour of human heart electrical nodes, propose a new bio-inspired algorithm inspired by the cardiac conduction system, and investigate its ability to solve benchmarks.

## 3. Algorithm Inspiration

In this section, the inspiration for the proposed method is discussed.

### 3.1 Cardiac Conduction System

The Cardiac Conduction System (CCS) is what makes the Heart go [26]. CCS is part of the Autonomic Nervous System (ANS). The ANS supplies, influences, and controls the system for regulating the function of internal organs [27]. However, the CCS is a group of heart muscle cells in the myocardium layer of the heart that causes blood to move throughout the body and produces and controls the heart rate through muscle cells that contract in waves [28]. It means whenever the first group of cells is stimulated. They have the effect of stimulating adjacent cells, and this chain of reaction continues until all cells are stimulated to enable the Heart to contract. Specifically, the electrical impulse is generated due to the distribution of ions (primarily calcium, sodium, potassium, and chloride) in cardiac cells [29]. The ions movement between in and out of the cell via dedicated ion channels in the cell membrane, which produce a wave of electricity flow across the cardiac tissues. The electrical wave propagates so that the ventricles and the atria contract in a harmonious rhythm.

Furthermore, CCS contains pacemaker cells with three unique characteristics: automaticity, conductivity, and contractility [28]. Automaticity is the ability of the cardiac tissues to produce an electrical impulse without any initiation factors, whereas conductivity is the capability to navigate the electrical impulse to the next neighbor cell. Contractility is the property of the cells to shorten the fibers in the Heart during receiving the electrical impulse [30]. In addition, the pacemaker cells with these mentioned unique characteristics coordinate the rhythmic relaxation and contraction of the Heart's chambers, called the cardiac cycle. The currents that exert a powerful effect on the heart cells are less than a millionth of an ampere [31]. The cardiac cycle has two main phases, diastole and systole, also called repolarisation and depolarisation. Diastole is the state of the Heart when the Heart's ventricles are relaxed and take longer than systole, which is about two-thirds of the cycle. Systole is the state of the Heart during which blood is pumped out of the ventricles and takes up the remaining one-third of the cycle [30]. Hence, the process of systole and diastole enables the Heart to pump blood and the volume of blood distributed by the Heart in one minute is known as the cardiac output, which the CCS produces. The significant elements of the CCS are the sinoatrial or sinus node (SA), the atrioventricular node (AV), the bundle of His (BoH), and the Purkinje fibers (PF) as shown in Figure 1.

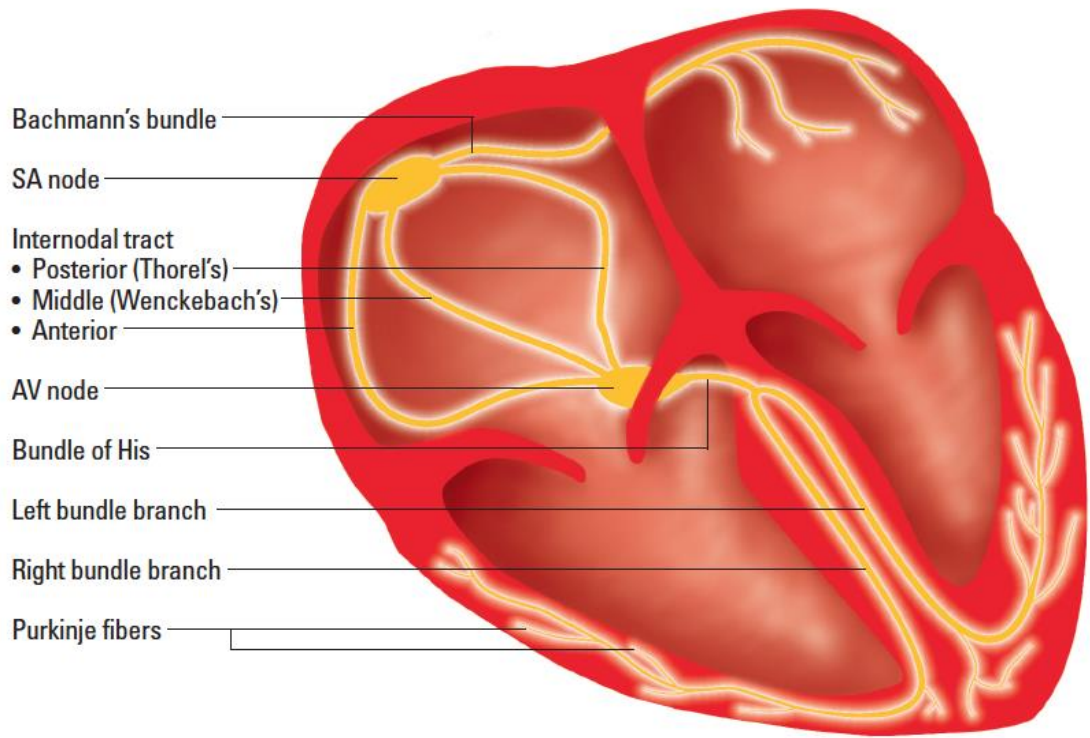

Figure 1: Cardiac Conduction System elements anatomy [28]

The CSS generates and controls heart rate among its elements through four main steps:

1. The first step (Sinoatrial node (SA node)): It is the dominant heart pacemaker that is located on the endocardial surface of the right atrium [28]. SA node generates impulse through the heart wall that enables atrial contraction [29].
2. Second Step (Atrioventricular node (AV node)): It can delay conduction between the ventricles and atria by a tenth of one second [26]. Hence, this node acts as a resistor to allow enough time for the contracting atria to transfer all blood to the ventricles before the ventricles contract (i.e., lower chambers of the Heart) [29].
3. The third step (Bundle of His): From the AV node, the impulse is passed to the bundle of His (BoH) and branches off the impulse into the left and right bundles with fewer charges [26].
4. Fourth step (Purkinje fibres): The impulse arrives at the Purkinje fibres (PF) located at the distal portions of the right and left bundle branches. The fibres across the surface of the ventricles cover the endocardium to the myocardium, which causes ventricular depolarisation [32].

Moreover, the SA node is the Heart's normal pacemaker that determines the Heart's rate, and any abnormality that occurs at this node the heart pace will face defeat in keeping the heart rate balanced [32]. As mentioned, the AV node acts as a resistor and slow conductions due to three main factors:

1. The diminished number of gap junctions between the successive cells in the conducting pathways.
2. Cells are made up of smaller diameter fibers.
3. Presence of Multiple sub-branches.

As a result, excellent resistance to the conduction of excitatory ions is produced from one conducting fibre cell to the next. Moreover, the electrical impulse is navigated in cardiac tissues with different conduction rates in a meter per second, as shown in Table 1. Hence, the time taken for the cardiac impulse to reach each node varies based on the speed of the conduction rate at each node. The time that requires the impulse travel from SA node to AV node is 0.03 seconds, between AV node and BoH is 0.13 seconds, then from BoH to PF is 0.03 seconds, and finally from PF to the endocardial and epicardia surface of ventricles is 0.03 seconds. However, any disturbance in the simulation sequence and timing of any node in the conduction system there will be covered by the next node. Otherwise, the heart rate will face a problem that is called conduction abnormality, which is discussed in the next section.

Table 1: Conduction rate in cardiac tissues [29]

| Conduction tissues | Conduction rate (M/S) |
|---|---|
| SA node | 0.05 |
| Atrial pathway | 1 |
| AV node | 0.05 |
| Bundle of His | 1 |
| Purkinje Fiber | 4 |
| Ventricular muscle | 1 |

**3.2 Clinical Significance of CCS**

The disturbance in the arrangement of the simulation of the specialised cardiac tissue lead to a rhythmic disturbance called arrhythmias, or conduction abnormality known as heart block [29]. In the case of rhythmic disturbance, the CCS has two built-in protection and safety mechanisms that will cover the failure of any node to navigate an impulse [28]. Thus, if the natural heart pacemaker (SA node) fails to fire, the AV node, called the secondary heart pacemaker, will generate an impulse between 45-60 beats per minute. The AV node takes the lead in generating an impulse under two conditions if the SA node is out of order and there is a blockage of conduction from the SA node. The second safety is ventricles that can generate impulses between 30 and 45 beats per minute if the SA node and AV node fail to fire [31]. To clarify more, the BoH generates an impulse between 40 and 45 beats per minute at the ventricles, and PF can generate its impulse between 30 and 40 beats per minute. As a result, these safety mechanisms enable the Heart to be one of the most precision scheduling organs that fight to the last minute to keep the balance for pumping blood to and from the body.

As mentioned, the Heart is innervated and affected by the ANS. This innervation sourced from the autonomic centre in the brainstem cannot simulate a heartbeat, but it can affect either decreasing or increasing the heartbeat rate [32]. Furthermore, the Heart's innervation consists of two parasympathetic and sympathetic components, collectively pointed to the coronary plexus in the brain. On the one hand, sympathetic innervation increases the permeability of the heart muscle fibres to calcium ions, increasing the force and heart contractions rate. On the other hand, parasympathetic innervation increases the permeability of heart muscle fibres to potassium ions, leading to a decrease in impulse generation and heart rate. Still, generally, it does not affect the force of the contractions. However, increasing or decreasing blood supply is required based on the situation during physical exertion or heart block. In this case, there will be natural methods for increasing heart power such as physical exercise and deep breathing. At the same time, chemical methods like eating drugs for activating cardiac nerves (i.e., sympathetic innervation) increase heart rate and vice versa [32]. Therefore, the heart rate range varies depending on the chemical and physical factors to regulate and provide the best performance suitable for the body. To sum up, the above behaviours of CCS and factors that affect CCS can be used to create a novel metaheuristics optimisation algorithm to find an optimal solution for real work and NP problems.

## 4. Artificial Cardiac Conduction System

From the biological facts mentioned in section 3, the behaviour of the CCS can be concluded in the following points:
1. Rhythmical electrical impulse generation
2. Built-in safety mechanism
3. Conduction rate
4. AV node delay
5. ANS effect (i.e., sympathetic and parasympathetic)

The above behaviours have been translated into a novel metaheuristic bio-inspired optimisation algorithm that we named ACCS. In addition, the details of the algorithm construction can be seen in (subsections 4.6 and 4.7), where we used CCS behaviours to construct the ACCS algorithm, which searches for the best solution in a loop using metaheuristic techniques to reach the optimal solution. The ACCS is proposed in an abstract high level in this paper that can be thought of as naturally little brain occurring in the learning metaheuristics system. However, one of the main characteristics of the CCS is the deep scheduling procedure that enables the Heart to have two built-in protection mechanisms to backup any failure node to generate or pass impulses through it. This biological feature can be described as a self-organisation feature, a spontaneous algorithmic approach adopted by the system that means without external interventions and adjusts its behaviour by a local agent inside the system. Hence, the CCS is an intelligent collective of nodes that strive to keep the heart pacing at the normal range, and the behaviour of this system is used to develop ACCS, which are detailed in the following subsections.

### 4.1 Rhythmical Electrical Impulse Generation

The CCS muscles can generate electrical impulses as they have pace characteristics to enable the heart to pump blood from the Heart to/from the body. The optimal heart rate value in the search space is not known a priori since the CCS is controlled via not one but four major nodes inside the myocardium layer of the Heart. This behaviour is applied in the ACCS algorithm assuming that the current best candidate solution is the target heartbeat value or close to the optimal heartbeat for an effective heart rating. CCS system remains the exploration process to optimise the heartbeat. After the best heart rate is defined, the other elements in the search area will then move and update towards the best search agent found, which is near the optimal solution. The ACCS algorithm starts to choose four random candidate solutions among all candidates over the search spaces (exploration) based on the problems to represent heart rate at each SA node, AV node, BoH node, and PF node. ACCS performs a random search for each agent in the search space and moves toward the best solution in a cooperative search with other agents across the search space. Therefore, R is used as a random factor with a value greater than 1 or less than $I_{HR}$ to force the conditions and search agent to move far away from a reference heart rate and explore the search space more efficiently. This behaviour is formulated in equation (4.1).

$$R_{SA,AV,BoH,PF} = randi\,[1\ I_{HR}] \tag{4.1}$$

where $I_{HR}$ is an index number for the maximum number of decision variables, and the same equation is used for all nodes conditions, the Heart is working in a very strategic manner to find the optimal factors that generate the most precise heartbeats.

For ACCS to seek an optimal solution, the best candidate heart rate solution is selected, and each heartbeat moves toward its previous best ($HR_{PB}$) solution concerning the current global best ($X^*(t)$) candidate heart rate solution using the equations (4.2) and (4.3).

$$HR_{PB}(i,t) = arg_{k=1,\ldots,t}\left[f\big(HR_i(k)\big)\right], i \in \{1,2,\ldots,N_{HR}\} \tag{4.2}$$

$$X^*(t) = arg_{i=1,\ldots,N_{HR}\ k=1,\ldots t}\left[f\big(HR_i(k)\big)\right], i \in \{1,2,\ldots,N_{HR}\} \tag{4.3}$$

where $i$ is denoted the heart rate index, $N_{HR}$ is the total number of heart rates, $t$ the current iteration number, $f$ the fitness function, and $HR$ is the heart rate value.

**4.2 Built-in Safety Mechanism**

The CCS has a deep scheduling procedure among four major nodes of the conduction system that enable the Heart to fight to the last point to keep pace to pump blood from the Heart to the body. Each node can cover the other node in case of the conduction blockage from one node to another that navigates the impulses through the Heart. However, the four randomly chosen candidate solutions in the rhythmical electrical impulse generation stage will be compared against the random number to determine whether the heart rate is normal or not at each node. This behaviour is represented by the following equation, which is inspired by the equation (4.4) used for calculating heart rate from EKG waveform.

$$Imp_{SA,AV,BoH,PF} = \frac{LB}{UP} * r \tag{4.4}$$

Where LB indicates the lower boundary of the problem, UP is the upper boundary of the problem, and r is a random number in [0, 1]. Here it is worth mentioning that (Imp) will be updated by a random number changing in each iteration. In addition, at each heart node, a heart rate is generated randomly using the same equation. Later, the heart rate will indicate if the heart node is at a normal pace or if the upcoming node needs to generate a heartbeat independently.

### 4.3 Conduction Rate

Another intercepting point is that sometimes dysfunction within the nodes themselves will cause that node to be either out of order or unable to generate impulse properly. Hence, this condition is obeyed for checking along with conduction blockage using either option to check if the Heart is in normal condition or not using a flag-like value via the equation (4.5).

$$flag_{T,SA,AV,BoH,PF} = \begin{cases} r_1, & [0, 0.5] \\ r_2, & [0.6, 1] \end{cases} \tag{4.5}$$

Where $r_1$ is flagged if there isn't any dysfunction at any of the 4 nodes, $r_2$ in case of having dysfunction at any of 4 nodes. Hence after passing through these two filters, the electrical impulses start to navigate through the heart muscles causing the Heart to start polarisation and depolarisation to force the pumping operation of the Heart. The current heart rate is updated considering the optimal heart rate obtained so far by this solution, which represents the exploitation mechanism of the algorithm to achieve better results in optimising the solutions. The heart rate at the SA node, BoH node, and PF node if the heart rate is in its normal state at these nodes are updated using equation (4.6).

$$\vec{X}(t+1) = \vec{R}.\left(\vec{X}(t) - r * X^*(t)\right) - \vec{X}(t) \tag{4.6}$$

Where $t$ indicates the current iteration, $\vec{R}$ is a random vector in [0,1], $r$ is a random number in [0,1], $X^*$ is the best heart rate solution obtained so far, $\vec{X}$ is the heart rate vector, is an element-by-element multiplication. Moreover, it is worth mentioning that $X^*$ should be updated if a better candidate solution is obtained in each iteration. Otherwise, if the heart rate at the SA node, BoH node, and PF node is not normal, then an additional factor will be added to the Heart to increase the efficiency of the blood pumping process, which is heart power. This behaviour is formulated as equation (4.7).

$$\vec{X}(t+1) = \vec{R}.\left(\vec{X}(t) - r * X^*(t)\right) - HP.\vec{X}(t) \tag{4.7}$$

Where $t$ indicates the current iteration, $\vec{R}$ is a random vector in [0,1], $r$ is a random number in [0,1], $X^*$ is the best heart rate solution obtained so far, $\vec{X}$ is the heart rate vector. is an element-by-element multiplication, and HP is the heart power factor. The HP is calculated and found in the basic of equation (4.8).

$$HP = 0.985 * C \tag{4.8}$$

where HP is linearly decreased from 0.985 to 0 throughout iterations (exploitation), C is a constant number (i.e., 0.999) chosen via trial and test to decrease the convergence time of the ACCS algorithm. In addition to equation (4.6) and equation (4.7), the impulse at the AV node is updated with the delay factor. The details and mathematical model for this update are in the following section.

### 4.4 AV Node Delay

In the CCS, the electrical impulse at the AV node faces a delay by one of the tenth seconds to allocate enough time to navigate blood efficiently from the heart's upper chambers to the heart's lower chambers (ventricles). Therefore, a delay factor should be added to each of equation (4.6) and equation (4.7) for updating the heart rate at the AV node whether the Heart is at a normal state. However, if the heart rate at the AV node is normal, then this updating behaviour is formulated in equation (4.9)

$$\vec{X}(t+1) = \vec{R}.\left(\vec{X}(t) - r * X^*(t)\right) - D_1.\vec{X}(t) \tag{4.9}$$

Where $t$ indicates the current iteration, $\vec{R}$ is a random vector in [0,1], $r$ is a random number in [0,1], $X^*$ is the best heart rate solution obtained so far, $\vec{X}$ is the heart rate vector. is an element-by-element multiplication, and $D_1$ is a normal delay factor which is a random number in [0.1,0.2]. If the heart rate is not normal at the AV node, then the random delay value will be increased to widen the search space and the considered heart power factor. This behaviour is illustrated in equation (4.10) to formulate nodes.

$$\vec{X}(t+1) = \vec{R}.\left(\vec{X}(t) - r * X^*(t)\right) - D_2.HP.\vec{X}(t) \tag{4.10}$$

Where $t$ indicates the current iteration, $\vec{R}$ is a random vector in [0,1], $r$ is a random number in [0,1], $X^*$ is the best heart rate solution obtained so far, $\vec{X}$ is the heart rate vector. is an element-by-element multiplication, HP is heart power factor generated using equation (8), and $D_2$ is a delay factor that is a random number in [0.2,0.3].

### 4.5 Autonomic Nervous System (ANS) Effect

ANS affects decreasing or increasing heart rate per minute. The sympathetic nervous system (SNS) and parasympathetic nervous system (PNS) are part of ANS that activates the fight or flight response. Hence, it affects the heart rate level of humans either by using physical or chemical factors that activate each SNS and PNS. This behaviour is achieved by adding the heart power ($HP$) factor in the equation (7) and equation (10). Note that the heart power acts as a random number to increase the search power over the search space for both equations to reach an optimal value in less time and prevent the case of local optimal. On the one hand, in the exploration phase, when the heart rate is not normal, the moving electrical impulses from one node to another are blocked. In such a case, this heart power action causes the candidate to search the problem space efficiently. Some candidate solutions keep repeating the movement at the close circle without any movement and activity in the search space without this action. However, this factor enforces the impulses to go away from this blockage and start a new search, increasing the chance to find global optima.

On the other hand, the ACCS is searching and updating the candidate solutions through four levels of checking for the same candidate solution in each iteration, which makes ACCS effective in exploiting the explored solutions. Hence, this assures the global convergence or the exploration of the algorithm to prevent the ACCS from the trap into a local solution. In addition to that, at later iterations of the ACCS algorithm, when passing through these four levels of updating and decreasing $HP$ value, will increase the exploitation capability of the algorithm. However, in ACCS proposed algorithm, the following steps in subsection 4.6 should be compiled and repeated based on the description above for the algorithm to find the optimal solution for the problem, and the pseudocode details are reported as shown in subsection 4.7. Therefore, in the next section, the benchmarking for the ACCS will be accomplished using standard test functions to demonstrate its performance.

**4.6  The Execution of the ACCS algorithm**

The flow chart (Figure 2) for the execution of the ACCS algorithm is presented below:

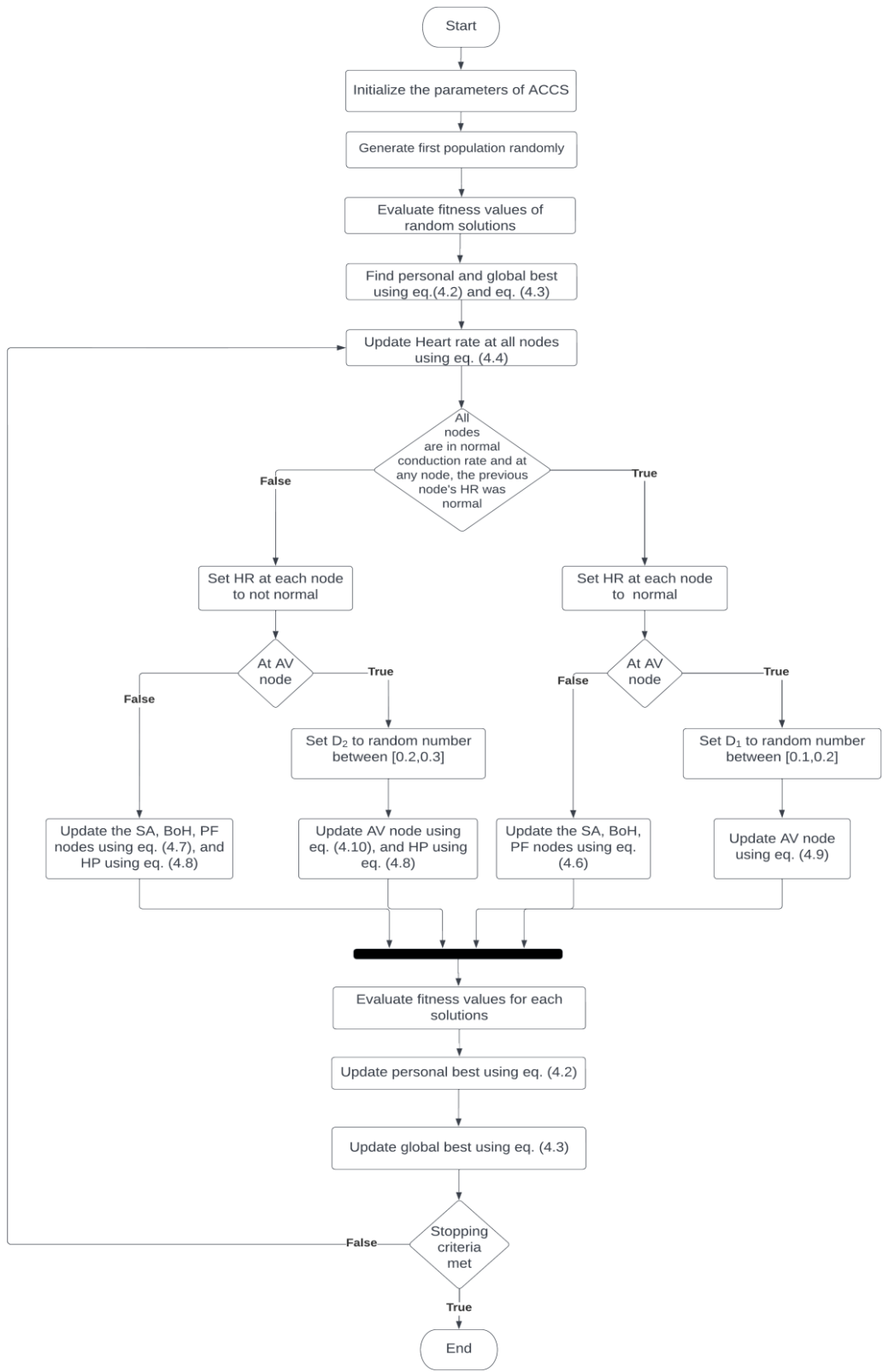

**Figure 2: The flow chart for the execution of the ACCS.**

### 4.7 Pseudocode for ACCS algorithm

The pseudo-code of the ACCS algorithm is presented in Figure 3 and illustrates the states of the solution to get the optimal values.

```
Initialization
        Initialize Population Members of Heartbeats Randomly
        Initialize ACCS's parameters Heart power, delay
repeat1 (until stopping criteria is met)
        Evaluate random solution (Finding fitness value using fitness function)
        Set the personal best, and global best
end repeat1
repeat2 (until stopping criteria is met)
        Updating heart impulse rate at all nodes using equation (4.4)
        Randomly set index to update the heart rates
        Set conduction rate flag for SA_node
                // for SA, BoH, PF nodes
                if1(Heart rate < nodes_impulse rate or conduction rate is not flagged)
                        Update Heart beat using equation (4.6)
                        Set Conduction rate flag to normal using equation (4.5)
                else if1
                        Update Heart beat using equation (4.7)
                        Set conduction rate flag to normal using equation (4.5)
                end if1
                // for AV node
                if2 (Heart rate < AV_node rate or conduction rate is not flagged)
                        Update Heart beat using equation (4.9)
                        Set conduction rate flag to normal using equation (4.5)
                        Set delay to $D_1$
                else if2
                        Update Heart beat using equation (4.10)
                        Set conduction rate flag to not normal using equation (4.5)
                        Set delay to $D_2$
                end if2
        Update the local best Heart rate using equation (4.2)
        Update the global best ($X^*$) if there is a better solution using equation (4.3)
end repeat2
return $X^*$ that holds the best found solution
```

Figure 3: ACCS Algorithm Pseudocode illustration

## 5. Experimental Performance Evaluation

To test the performance of the ACCS algorithm, several standard benchmark functions that have been conducted to evaluate the performance of the other existed algorithm in the literature are used. Additionally, the experimentally obtained results will be demonstrated, and these results are compared to five other well-known optimisation algorithms, which are Whale Optimization Algorithm (WOA), Particle Swarm Optimization (PSO), Gravitational Search Algorithm (GSA), Differential Evolution (DE), and Fast Evolutionary Programming (FEP). It is worth mentioning that the outcome developed in this study for ACCS was tested by solving 19 mathematical optimisation problems. All 19 solved problems are classical benchmark test functions tested on all five mentioned algorithms along with ACCS [10]. Note that *dim* indicates the dimension, *the range* is the lower and upper boundary of the standard test functions' search space, and $F_{min}$ is the minimum objective function value in Table (2, 3, and 4). Many researchers in their studies have used these classical standard test functions to evaluate their algorithm performance [3-14] and many others. However, the characteristics of benchmark functions, numerical experiments, parameter settings, and performance analysis of the ACCS are evaluated, and the result will be demonstrated in the following sections.

**5.1 Benchmark Functions**

**5.1.1. Classical Test Functions**

Generally, the ACCS algorithm is tested using two sets of classical test functions to determine the algorithm's performance. The benchmark functions can be divided into several categories based on different characteristics: unimodal test functions, multimodal test functions, and fixed-dimensions multimodal test functions [34]. Figures (4, 5, and 6) represent the typical 2D version plots of the benchmark mathematical test functions for some mathematical test cases conducted in this paper. However, in the benchmark test functions, whenever the number of dimensions gets high, the local minima increase exponentially, affecting the complexity of the problem to find the optimal value [35]. Therefore, each set of the considered test functions is used to compare the particular perspective of the ACCS algorithm. For instance, unimodal benchmark test functions as indicated by their name that they have a single optimum, and they are used to check the performance in terms of exploitation level and convergence of the algorithms. At the same time, multimodal benchmark test functions and fixed-dimensions multimodal test functions, as their name might imply that these types of test functions have multi-optimal solutions. They are used for testing the exploration and local optima avoidance of the algorithm. In addition to the fact that, in the multimodal algorithms, there are more than several optimum solutions composed of one global optimum solution and several local optimum solutions. Hence, the ACCS must avoid local optimum and converge toward a globally optimum solution.

Table 2: Description of unimodal benchmark test functions

| Function | dim | Range | $F_{min}$ |
|---|---|---|---|
| $f_1(x) = \sum_{i=1}^{n} x_i^2$ | 30 | [-100,100] | 0 |
| $f_2(x) = \sum_{i=1}^{n} |x_i| + \prod_{i=1}^{n} |x_i|$ | 30 | [-10,10] | 0 |
| $f_3(x) = \sum_{i=1}^{n} (\sum_{j-1}^{i} x_j)^2$ | 30 | [-100,100] | 0 |
| $f_4(x) = max_i\{|x_i|, 1 \leq |i| \leq n\}$ | 30 | [-100,100] | 0 |
| $f_5(x) = \sum_{i=1}^{n-1} [100(x_{i+1} - x_i^2)^2 + (x_i - 1)^2]$ | 30 | [-30,30] | 0 |
| $f_6(x) = \sum_{i=1}^{n} ([x_i + 0.5])^2$ | 30 | [-100,100] | 0 |
| $f_7(x) = \sum_{i=1}^{n} ix_i^4 + random[0,1]$ | 30 | [-1.28,1.28] | 0 |

Table 3: Description of multimodal benchmark test functions

| Function | dim | Range | $F_{min}$ |
|---|---|---|---|
| $f_8(x) = \sum_{i=1}^{n} -x_i \sin(\sqrt{|x_i|})$ | 30 | [-500,500] | -418.9829x5 |
| $f_9(x) = \sum_{i=1}^{n} [x_i^2 - 10\cos(2\pi x_i) + 10]$ | 30 | [-5.12,5.12] | 0 |
| $f_{10}(x) = -20 \exp\left(-0.2\sqrt{\frac{1}{n}\sum_{i=1}^{n} x_i^2}\right) - \exp(\frac{1}{n}\sum_{i=1}^{n} \cos 2\pi x_i + 20 + e)$ | 30 | [-32,32] | 0 |
| $f_{11}(x) = \frac{1}{4000} \sum_{i=1}^{n} x_i^2 - \prod_{i=1}^{n-1} \cos\left(\frac{x_i}{\sqrt{i}}\right) + 1$ | 30 | [-600,600] | 0 |
| $f_{12}(x) = \frac{\pi}{n}\{10\sin(\pi y_1) + \sum_{i=1}^{n-1}(y_i - 1)^2[1 + 10sin^2(\pi y_{i+1})] + (y_n - 1)^2\}$ $+ \sum_{i=1}^{n} u(x_i, 10, 100, 4)$ $y_i = 1 + \frac{x_i + 1}{4} \quad u(x_i, a, k, m) = \begin{cases} k(x_i - a)^m & x_i > a \\ 0 & -a < x_i < a \\ k(-x_i - a)^m & x_i < -a \end{cases}$ | 30 | [-50,50] | 0 |
| $f_{13}(x) = 0.1\{sin^2(3\pi x_1)$ $+ \sum_{i=1}^{n}(x_i - 1)^2[1 + sin^2(3\pi x_i + 1)] + x_n$ $- 1)^2[1 + sin^2(2\pi x_n)]\} + \sum_{i=1}^{n} u(x_i, 5, 100, 4)$ | 30 | [-50,50] | 0 |

**Table 4: Description of fixed-dimension multimodal benchmark test functions**

| Function | dim | Range | $F_{min}$ |
|---|---|---|---|
| $f_{14}(x) = (\frac{1}{500} + \sum_{j=1}^{25} \frac{1}{j + \sum_{i=1}^{2}(x_i - a_{ij})^6})^{-1}$ | 2 | [-65,65] | 1 |
| $f_{15}(x) = \sum_{i=1}^{11} \left[a_i - \frac{x_1(b_i^2 + b_i x_2)}{b_i^2 + b_i x_3 + x_4}\right]^2$ | 4 | [-5,5] | 0.00030 |
| $f_{16}(x) = 4x_1^2 - 2.1x_1^4 + \frac{1}{3}x_1^6 - x_1 x_2 - 4x_2^2 + 4x_2^4$ | 2 | [-5,5] | -1.0316 |
| $f_{17}(x) = \left(x_2 - \frac{5.1}{4\pi^2}x_1^2 + \frac{5}{\pi}x_1 - 6\right)^2 + 10(1 - \frac{1}{8\pi})\cos x_1 + 10$ | 2 | [-5,5] | 0.398 |
| $f_{18}(x) = [1 + (x_1 + x_2 + 1)^2(19 - 14x_1 + 3x_1^2 - 14x_2 + 6x_1 x_2 + 3x_2^2)]$ $\times [30 + (2x_1 - 3x_2)^2 \times (18 - 32x_1 + 12x_1^2 + 48x_2 - 36x_1 x_2 + 27x_2^2)]$ | 2 | [-2,2] | 3 |
| $f_{19}(x) = -\sum_{i=1}^{4} c_i \exp\left(-\sum_{j=1}^{3} a_{ij}(x_j - p_{ij})^2\right)$ | 3 | [1,3] | -3.86 |

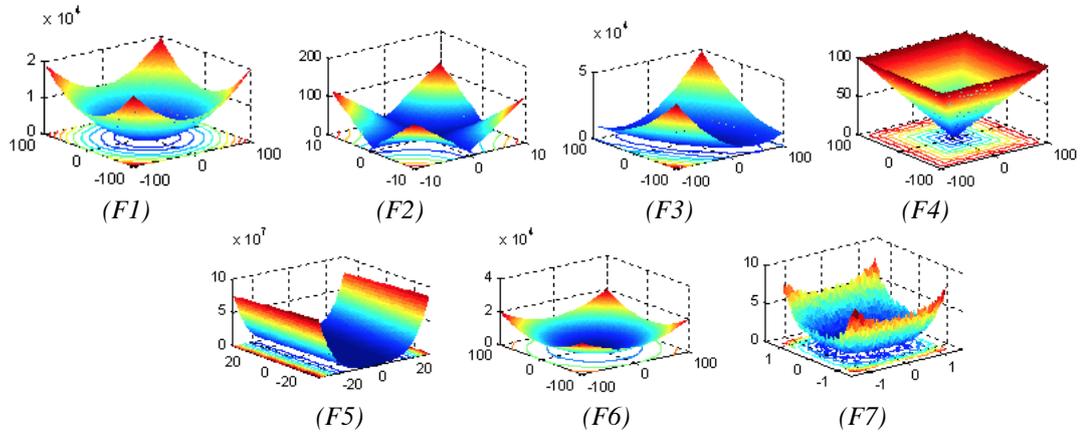

**Figure 4: 2D representations of unimodal benchmark functions**

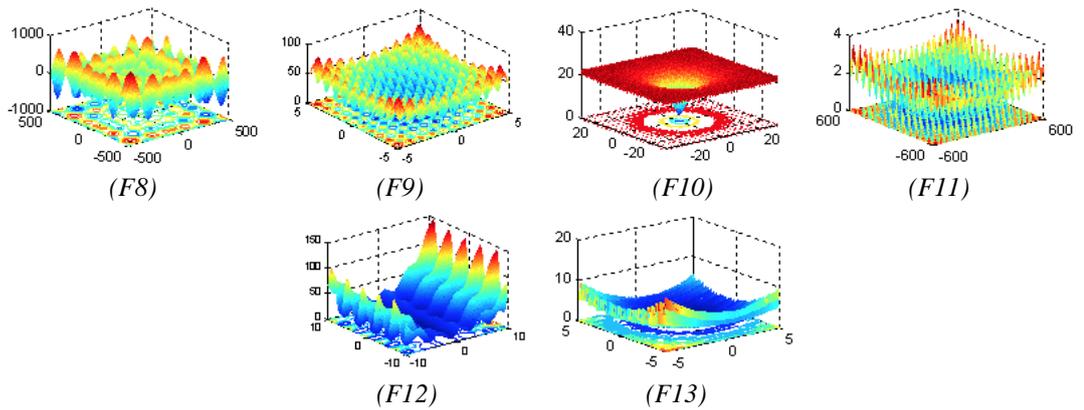

**Figure 5: 2D representations of multimodal benchmark functions**

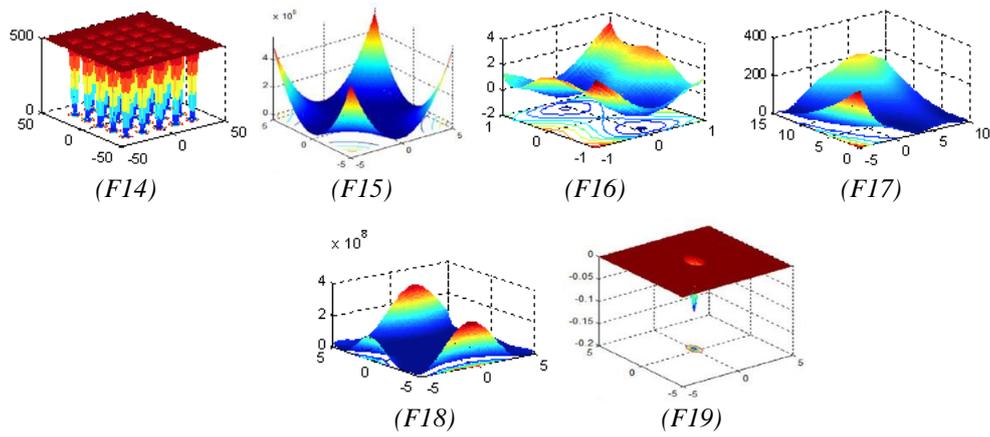

Figure 6: 2D representations of fixed-dimension multimodal benchmark functions

### 5.1.2. CEC-C06 2019 Benchmark Test Functions

In certain conditions, securing precision in the solution holds superior importance than achieving rapid outcomes. Additionally, various individuals have the capability to enhance an algorithm and iterate its execution. Users arrange the algorithm that best aligns with their specific requirements, irrespective of the time asset involved. Within this contemporary benchmark compilation, the CEC-2019 conference introduced ten test functions [36]. The performance of these functions has been assessed using the ACSS algorithm. Referred to as "The 100-Digit Challenge," these test functions are designed for utilization in annual optimization competitions, as illustrated in Table (5). The functions unveiled during CEC-2019 have gained widespread popularity and are recognized as state-of-the-art benchmarks for appraising algorithmic performance in tackling real-world issues.

Table 5: CEC-2019 benchmarks "the 100-digit challenge" (Brest et al., 2019)

| $f$ | Functions | Dim | Range | $f_{min}$ |
|---|---|---|---|---|
| CEC01 | Storn's Chebyshev Polynomial Fitting Problem | 9 | [-8192, 8192] | 1 |
| CEC02 | Inverse Hilbert Matrix Problem | 16 | [-16384, 16384] | 1 |
| CEC03 | Lennard-Jones Minimum Energy Cluster | 18 | [-4,4] | 1 |
| CEC04 | Rastrigin's Function | 10 | [-100, 100] | 1 |
| CEC05 | Griewangk's Function | 10 | [-100, 100] | 1 |
| CEC06 | Weierstrass Function | 10 | [-100, 100] | 1 |
| CEC07 | Modified Schwefel's Function | 10 | [-100, 100] | 1 |
| CEC08 | Expanded Schaffer's F6 Function | 10 | [-100, 100] | 1 |
| CEC09 | Happy Cat Function | 10 | [-100, 100] | 1 |
| CEC10 | Ackley Function | 10 | [-100, 100] | 1 |

### 5.2 Numerical Experiment

The ACCS algorithm is coded in the MATLAB R2017b and implemented on the MacBook Pro (Retina, 15-inch, Late 2013), Graphic Intel Iris Pro 1536 MB, Processor 2.6 GHz Intel Core i7, Memory 16 GB 1600 MHz DDR3. On the contrary, the numerical results of WOA, PSO, GSA, DE, and FEP algorithms are taken from [10].

### 5.3 Parameter Setting

In ACCS, WOA, PSO, GSA, DE, FEP, DA, Leo, and ANA algorithms, the parameters have been set as follows:

1. The number of search agents (solution candidate) was equal to 30 except in fixed-dimensions multimodal functions, which is according to the range of each function as shown in Table (5.1-5.3);
2. Population size was equal to 30;
3. The maximum number of iterations was equal to 500;
4. Heart power factor.

### 5.4 Analysis and Discussion

In this section, the ACCS algorithm's effectiveness has been evaluated and determined performance against other recently published algorithms. Usually, the testing is accomplished by solving a set of standard benchmark problems [33]. Notably, 19 classical test functions have been used to compare and evaluate the performance of the ACCS with other recent metaheuristics algorithms. These classical test functions are categorised into three types, which are unimodal ($F_1 - F_7$), multimodal ($F_8 - F_{13}$), fixed-dimension multimodal ($F_{14} - F_{19}$), and the mathematical equation of these functions are shown in Table (5.1-5.3) respectively. Additionally, Table 4 presents the mathematical of cluster scripts based on the CEC-C06 2019 Test Functions.

The ACCS, WOA, PSO, GSA, DE, FEP DA, Leo, and ANA variants were compiled 30 times on each benchmark test function with setting up the parameters, as mentioned in section 5.3. In each test, the algorithm was enabled to search for the best optimum solution via running the algorithm for 500 iterations. In addition, the mean and standard deviation of the optimal solutions for each function in these 30 runs were calculated, and then, they were compared to the mean and standard deviation of the other algorithms reported in Table (5,6,7 and 8). The verification of the performance of the ACCS in terms of exploitation and exploration capability against the other algorithms is discussed in the following subsections.

#### 5.4.1 Evaluation of Exploitation Capability (F1-F7)

As previously discussed, functions F1-F7 represent unimodal functions since they have only one global optimal value, and these functions allow testing the exploitation capability of the metaheuristics algorithms. Table (5.4) shows that the ACCS gives highly competitive solutions compared with WOA, PSO, GSA, DE, and FEP. Therefore, ACCS has a high rate of exploitation capability and is very competitive with other metaheuristics algorithms. In particular, the F1 to F4 and F7 results showed that ACCS is the most efficient optimiser among other compared algorithms; however, the F5 results represented that the ACCS is better than PSO and GSA algorithms while WOA, DE, and FEP algorithms were better than ACCS algorithm. The ACCS in F6 showed poor performance than the other five algorithms. Thus, the ACCS algorithm provides excellent exploitation due to the behaviour of the conduction systems mathematical model, which passes through four different conditions and then updates the candidate solutions four times in each iteration.

### 5.4.2 Evaluation of Exploration Capability (F8-F19)

Unlike unimodal standard test functions, multimodal test functions have many local optima whose number is controlled by the number of design variables (i.e., problem size). Hence, the number of local optima increases exponentially with increasing the number of design variables. Therefore, this test function can be useful for evaluating the exploration capability of the ACCS algorithm. Furthermore, the statistical results of the proposed ACCS algorithms on multimodal test functions are reported in Table (6 and 7). It can be observed that the F8, F9, F10, and F11 results presented that the ACCS is the most efficient optimiser than the other compared algorithms. In contrast, in F12 and F13 ACCS provided a better experimental, numerical result than GSA but poor performance than the other algorithms. However, in F14 to F19, ACCS results showed poor performance than other algorithms as they have fixed dimensions. Hence, in general, the results confirm that the ACCS algorithm has a good exploration capability. In addition to the fact that the integrated mechanisms of the exploration are represented by some factors inside the ACCS calculation, for instance, heart power and delay that make sure to search the algorithm can explore the search space extensively.

Generally speaking, ACCS is the first algorithm that uses human heart behaviour to model mathematical equations, then produces a novel metaheuristic optimisation algorithm. ACCS explains and shows an intense scheduling procedure existing inside the human heart that controls the repolarisation and depolarisation of the Heart. Unlike most algorithms, ACCS implemented the cardiac conduction process in high-level procedures that matched theoretical and technical details. The factors that affect ACCS are used in the algorithm's development to increase the algorithm's exploration capability. ACCS is easily tuned and implemented for various real-world problems due to the simple concept structure of the algorithm. ACCS can update the candidate solutions four times in a single iteration due to the behaviour of the four nodes of the ACCS. To sum up, all the mentioned experimental and numerical results assert that the ACCS is very efficient in exploiting and exploring the test problems compared to WOA, PSO, GSA, DE, FEP algorithms in terms of computational efforts and quality of the results. However, the results of the ACCS were poor performance in the fixed-dimensions multimodal test functions. On the contrary, in unimodal and multimodal test functions, the results are highly competitive, and in most of them, it was the most efficient optimizer.

**Table 6: The average of unimodal, multimodal, and fixed-dimension multimodal classical benchmark functions**

| F | ACCS | WOA | PSO | GSA | DE | FEP |
| --- | --- | --- | --- | --- | --- | --- |
|  | Avg. | Avg. | Avg. | Avg. | Avg. | Avg. |
| F1 | 0 | 1.41E-30 | 0.000136 | 2.53E-16 | 8.20E-14 | 5.70E-04 |
| F2 | 0 | 1.06E-21 | 0.042144 | 0.055655 | 1.50E-09 | 0.0081 |
| F3 | 0 | 5.39E-07 | 70.12562 | 896.5347 | 6.80E-11 | 0.016 |
| F4 | 0 | 0.072581 | 1.086481 | 7.35487 | 0 | 0.3 |

| F | | | | | | |
|---|---|---|---|---|---|---|
| F5 | 28.9653 | 27.86558 | 96.71832 | 67.54309 | 0 | 5.06 |
| F6 | 7.0137405 | 3.116266 | 0.000102 | 2.50E-16 | 0 | 0 |
| F7 | 4.82E-05 | 0.001425 | 0.122854 | 0.089441 | 0.00463 | 0.1415 |
| F8 | -2465.5153 | -5080.76 | -4841.29 | -2821.07 | -11080.1 | -12554.5 |
| F9 | 0 | 0 | 46.70423 | 25.96841 | 69.2 | 0.046 |
| F10 | 8.88E-16 | 7.4043 | 0.276015 | 0.062087 | 9.70E-08 | 0.018 |
| F11 | 0 | 0.000289 | 0.009215 | 27.70154 | 0 | 0.016 |
| F12 | 1.3991 | 0.339676 | 0.006917 | 1.799617 | 7.90E-15 | 9.20E-06 |
| F13 | 2.9919 | 1.889015 | 0.006675 | 8.899084 | 5.10E-14 | 0.00016 |
| F14 | 10.6834 | 2.111973 | 3.627168 | 5.859838 | 0.998004 | 1.22 |
| F15 | 0.036683 | 0.000572 | 0.000577 | 0.003673 | 4.50E-14 | 0.0005 |
| F16 | -0.92105 | -1.03163 | -1.03163 | -1.03163 | -1.03163 | -1.03 |
| F17 | 1.9425 | 0.397914 | 0.397887 | 0.397887 | 0.397887 | 0.398 |
| F18 | 34.0941 | 3 | 3 | 3 | 3 | 3.02 |
| F19 | -3.1951 | -3.85616 | -3.86278 | -3.86278 | N/A | -3.86 |

**Table 7: The standard deviation of unimodal, multimodal, and fixed-dimension multimodal classical benchmark functions**

| F | ACCS | WOA | PSO | GSA | DE | FEP |
|---|---|---|---|---|---|---|
| | Std | Std | Std | Std | Std | Std |
| F1 | 0 | 4.91E-30 | 0.000202 | 9.67E-17 | 5.90E-14 | 0.00013 |
| F2 | 0 | 2.39E-21 | 0.045421 | 0.194074 | 9.90E-10 | 0.00077 |
| F3 | 0 | 2.93E-06 | 22.11924 | 318.9559 | 7.40E-11 | 0.014 |
| F4 | 0 | 0.39747 | 0.317039 | 1.741452 | 0 | 0.5 |
| F5 | 0.037608 | 0.763626 | 60.11559 | 62.22534 | 0 | 5.87 |
| F6 | 0.24026299 | 0.532429 | 8.28E-05 | 1.74E-16 | 0 | 0 |
| F7 | 3.84E-05 | 0.001149 | 0.044957 | 0.04339 | 0.0012 | 0.3522 |
| F8 | 525.5786 | 695.7968 | 1152.814 | 493.0375 | 574.7 | 52.6 |
| F9 | 0 | 0 | 11.62938 | 7.470068 | 38.8 | 0.012 |
| F10 | 0 | 9.897572 | 0.50901 | 0.23628 | 4.20E-08 | 0.0021 |
| F11 | 0 | 0.001586 | 0.007724 | 5.040343 | 0 | 0.022 |
| F12 | 0.17402 | 0.214864 | 0.026301 | 0.95114 | 8.00E-15 | 3.60E-06 |
| F13 | 0.015473 | 0.266088 | 0.008907 | 7.126241 | 4.80E-14 | 0.000073 |
| F14 | 3.3693 | 2.498594 | 2.560828 | 3.831299 | 3.30E-16 | 0.56 |
| F15 | 0.029559 | 0.000324 | 0.000222 | 0.001647 | 0.00033 | 0.00032 |

| F | | | | | | |
|---|---|---|---|---|---|---|
| F16 | 0.17111 | 4.20E-07 | 6.25E-16 | 4.88E-16 | 3.10E-13 | 4.90E-07 |
| F17 | 1.5656 | 2.70E-05 | 0 | 0 | 9.90E-09 | 1.50E-07 |
| F18 | 31.4115 | 4.22E-15 | 1.33E-15 | 4.17E-15 | 2.00E-15 | 0.11 |
| F19 | 0.4259 | 0.002706 | 2.58E-07 | 2.29E-15 | N/A | 0.000014 |

### 5.4.3. Evaluation CEC-C06 2019 (CEC01- CEC10)

As previously noted, this algorithm undergoes benchmarking using the CEC-C06 2019 Benchmark Test Functions. This process gauges the algorithm's capacity for exploitation and exploration, with results validated through a comparative study involving the DA, WOA, PSO, Leo, and ANA.

Table 8 underscores algorithm performance diversity across various test functions, with the basic ACCS function demonstrating notable consistency, implying stable performance across algorithms. This insight offers valuable perspectives on algorithms' general approaches to optimization challenges. Also, Table 9 utilizes standard deviations to gauge performance variability among algorithms. Particularly for the basic ACCS function, a low standard deviation signifies consistent performance across algorithms, indicating reliability in addressing this function. This consistent trend suggests algorithms consistently exhibit either strong or weak performance, showcasing minimal variation in their approach to the basic ACCS function.

Also, in both Table 8 and Table 9, the analysis of CEC05 reveals variations in the average performance of algorithms across diverse test functions. Specifically focusing on the basic ACCS result for CEC05, a consistent trend in the average performance of algorithms is observed. Upon scrutiny of the standard deviation for CEC05, the variability in performance among different algorithms becomes outward. Notably, for the basic ACCS function (CEC05), standard deviations exhibit relatively low values across algorithms, indicative of consistent performance. This observation suggests that algorithms consistently exhibit either proficient or deficient performance, with diminished variability in their execution when addressing the basic ACCS function.

Table 8: The average result of CEC-C06 2019 benchmark

| F | ACCS | DA | WOA | PSO | LEO | ANA |
|---|---|---|---|---|---|---|
| | Avg. | Avg. | Avg. | Avg. | Avg. | Avg. |
| CEC01 | 1.54E+05 | 5.43E+10 | 4.11E+10 | 1.47E+12 | 7294147266 | 8.91E+09 |
| CEC02 | 19.20848 | 78.0368 | 17.3495 | 15183.91 | 17.47763 | 4 |
| CEC03 | 12.70279 | 13.7026 | 13.7024 | 12.70280 | 12.70311 | 13.70240422 |
| CEC04 | 25070.53 | 344.3561 | 394.6754 | 16.80078 | 69.8652733 | 38.50887822 |
| CEC05 | 6.13997 | 2.5572 | 2.73420 | 1.138265 | 2.76024667 | 1.224598709 |
| CEC06 | 11.32554 | 9.8955 | 10.7085 | 9.305312 | 3.01982 | 10.456789 |
| CEC07 | 1492.699 | 578.9531 | 490.6843 | 160.6863 | 195.558303 | 116.5962143 |
| CEC08 | 6.768143 | 6.8734 | 6.9090 | 5.224137 | 5.06228333 | 5.472814997 |
| CEC09 | 3362.285 | 6.0467 | 5.9371 | 2.373279 | 3.26147 | 2.000963996 |
| CEC10 | 20.28013 | 21.2604 | 21.2761 | 20.28063 | 20.0123867 | 2.718281828 |

Table 9: The standard deviation of CEC-C06 2019 benchmark

| F | ACCS STD | DA STD | WOA STD | PSO STD | LEO STD | ANA STD |
|---|---|---|---|---|---|---|
| CEC01 | 84381.84 | 6.69E+10 | 5.42E+10 | 1.32E+12 | 5767198154 | 8.97E+09 |
| CEC02 | 0.323054 | 87.7888 | 0.0045 | 3729.553 | 0.09810875 | 2.87E-14 |
| CEC03 | 0.00155 | 0.0007 | 0 | 9.03E-15 | 0.00088954 | 2.01E-11 |
| CEC04 | 8443.25 | 414.0982 | 248.5627 | 8.19908 | 23.7808956 | 10.07245727 |
| CEC05 | 1.104676 | 0.3245 | 0.2917 | 0.08939 | 0.43275426 | 0.114632394 |
| CEC06 | 0.687164 | 1.6404 | 1.0325 | 9.30531 | 0.75595651 | 1.5567 |
| CEC07 | 284.8011 | 329.3983 | 194.8318 | 160.6863 | 236.53515 | 8.825046006 |
| CEC08 | 0.299269 | 0.5015 | 0.4269 | 0.786761 | 0.45975194 | 0.429461877 |
| CEC09 | 1003.938 | 2.871 | 1.6566 | 0.018437 | 0.74449295 | 0.00341781 |
| CEC10 | 0.222432 | 0.1715 | 0.1111 | 0.128531 | 0.0285509 | 4.44E-16 |

## 5.5 Classical Benchmark Comparative Study

As shown in Table (5.5), the global average experimental numerical results of the ACCS is 3.158 on a scale of 1 to 6, where the consideration for the scale is to be one of the best optimiser algorithms and six the worst optimizer algorithm. It is worth mentioning that the ranking of the ACCS is ranked based on the other five compared algorithms. In this manner, the ACCS algorithm was ranked first nine times, fourth once, fifth four times, and sixth five times out of nineteenth. However, the overall average of the ranking by the kind of benchmark test functions is as follows:

- Unimodal test functions (F1-F7): 2.143
- Multimodal test functions (F8-F13): 2.333
- Fixed-dimensions test functions (F14-F19): 5.667

The above results show that the ACCS algorithm used has a good balance between exploitation and exploration that assists this algorithm in searching and finding the global optima efficiently. Therefore, the ACCS is ranked third among all six algorithms and nineteenth test functions.

Table 5. 1 Ranking of the optimisation performance for ACCS, WOA, PSO, GSA, DE, and FEP to the classical benchmark test function

| F | 1st | 2nd | 3rd | 4th | 5th | 6th | Rank | Subtotal |
|---|---|---|---|---|---|---|---|---|
| F1 | ACCS | WOA | GSA | DE | FEP | PSO | 1 | |

| | | | | | | | |
|---|---|---|---|---|---|---|---|
| F2 | ACCS | WOA | DE | FEP | PSO | GSA | 1 | |
| F3 | ACCS | DE | WOA | FEP | PSO | GSA | 1 | |
| F4 | ACCS, DE | WOA | FEP | PSO | GSA | | 1 | |
| F5 | DE | FEP | WOA | ACCS | GSA | PSO | 4 | |
| F6 | FEP, DE | GSA | PSO | WOA | | ACCS | 6 | |
| F7 | ACCS | WOA | DE | GSA | PSO | FEP | 1 | 15 |
| F8 | ACCS | GSA | PSO | WOA | DE | FEP | 1 | |
| F9 | ACCS, WOA | FEP | GSA | PSO | DE | | 1 | |
| F10 | ACCS | DE | FEP | GSA | PSO | WOA | 1 | |
| F11 | ACCS, DE | WOA | PSO | FEP | GSA | | 1 | |
| F12 | DE | FEP | PSO | WOA | ACCS | GSA | 5 | |
| F13 | DE | FEP | PSO | WOA | ACCS | GSA | 5 | 14 |
| F14 | DE | FEP | WOA | PSO | GSA | ACCS | 6 | |
| F15 | FEP | WOA | PSO | GSA | ACCS | DE | 5 | |
| F16 | WOA, PS0, GSA, DE | FEP | | | | ACCS | 6 | |
| F17 | FEP | WOA | PSO, GSA, DE | | | ACCS | 6 | |
| F18 | WOA, PSO, GSA, DE | FEP | | | | ACCS | 6 | |
| F19 | FEP | PSO, GSA | WOA | | ACCS | DE | 5 | 34 |

| | | |
|---|---|---|
| | Total: | 63 |
| | Overall Rank: | 63/19=3.158 |
| | F1-F7: | 15/7=2.143 |
| | F8-F13: | 14/6=2.333 |
| | F14-F19: | 34/6=5.667 |

## 6. Conclusion

This work proposed a novel bio-inspired metaheuristic algorithm inspired by the human heart function. The proposed method mimicked the electric generation and control behaviour of the Heart. Nineteen test functions were employed to benchmark the proposed algorithm's performance in terms of exploration and exploitation. The results showed that ACCS could provide highly competitive results compared to well-known metaheuristics such as WOA, PSO, GSA, DE, and FEP. First, the results on the unimodal functions demonstrated the superior exploitation of the ACCS algorithm. Second, the results on multimodal functions confirmed the exploration ability of ACCS. Finally, the results on both types of functions confirmed that ACCS has a high convergence speed in fewer iterations and computational time.

For future work, we are going to develop a multi-objective version of the ACCS algorithm, study parameters in more detail to improve the results toward better performance and adapt parameters to a much more comprehensive range of applications, test a hybrid ACCS algorithm on a vast landscape global optimisation, and present a real-world application of the ACCS algorithm in the various fields.

**Conflict of interest:** The authors declare no conflict of interest to any party.

**Ethical Approval:** The manuscript is conducted within the ethical manner advised by the targeted journal.

**Consent to Participate:** Not applicable

**Consent to Publish:** The research is scientifically consented to be published.

**Funding:** The research received no funds.

**Competing Interests:** The authors declare no conflict of interest.

**Availability of data and materials:** Data can be shared upon request from the corresponding author.

**Acknowledgment:** None.

**References**

[1] Amarel, S. (1985). The History of Artificial Intelligence at Rutgers. *AI Magazine*, *6*(3), 192. https://doi.org/10.1609/aimag.v6i3.499

[2] C. Li, H. Peng, A. Xu, and S. Wang, "Immune System and Artificial Immune System Application," World Congress on Medical Physics and Biomedical Engineering 2006. Springer Berlin Heidelberg, pp. 477–480. doi: 10.1007/978-3-540-36841-0_128.

[3] M. Dorigo, M. Birattari, and T. Stutzle, "Ant colony optimisation," Computational Intelligence Magazine, IEEE, vol. 1, pp. 28-39, 2006.

[4] J. Kennedy and R. Eberhart, "Particle swarm optimisation," in *Neural Networks, 1995. Proceedings., IEEE International Conference on*, 1995, pp. 1942-1948.

[5] Mirjalili, S. (2016). Dragonfly algorithm: a new meta-heuristic optimization technique for solving single-objective, discrete, and multi-objective problems. Neural computing and applications, 27, 1053-1073.

[6] Mirjalili, S., Mirjalili, S. M. and Lewis, A. (2014) 'Grey Wolf Optimizer', *Advances in Engineering Software*, 69, pp. 46–61. doi: 10.1016/j.advengsoft.2013.12.007.

[7] S. Bandyopadhyay and R. Bhattacharya, "Discrete and Continuous Simulation", 2014. Available: 10.1201/b17127 [Accessed 20 August 2022].

[8] D. H. Wolpert and W. G. Macready, "No free lunch theorems for optimisation," *Evolutionary Computation, IEEE Transactions on,* vol. 1, pp. 67-82, 1997.

[9] F. Glover, "Future paths for integer programming and links to artificial intelligence," Computers & Operations Research, vol. 13, no. 5. Elsevier BV, pp. 533–549, Jan. 1986. doi: 10.1016/0305-0548(86)90048-1.

[10] S. Mirjalili and A. Lewis, "The Whale Optimization Algorithm," Advances in Engineering Software, vol. 95. Elsevier BV, pp. 51–67, May 2016. doi: 10.1016/j.advengsoft.2016.01.008.

[11] J. H. Dorigo, "Genetic algorithms," *Scientific american,* vol. 267, pp. 66-72, 1992.

[12] B. Basturk and D. Karaboga, "An artificial bee colony (ABC) algorithm for numeric function optimisation," in *IEEE swarm intelligence symposium*, 2006, pp. 12-14.

[13] Aladdin, A. M., & Rashid, T. A. (2023). Leo: Lagrange Elementary Optimization. *arXiv preprint arXiv:2304.05346.*

[14] Hama Rashid, D. N., Rashid, T. A., & Mirjalili, S. (2021). ANA: Ant nesting algorithm for optimizing real-world problems. *Mathematics*, *9*(23), 3111.

[15] X. S. Yang, *Nature-inspired metaheuristic algorithms*: Luniver Press, 2011.

[16] S. Kirkpatrick, D. G. Jr., and M. P. Vecchi, "Optimisation by simulated annealing," *science,* vol. 220, pp. 671-680, 1983.

[17] Zong Woo Geem, Joong Hoon Kim and G. Loganathan, "A New Heuristic Optimization Algorithm: Harmony Search", *SIMULATION*, vol. 76, no. 2, pp. 60-68, 2001. Available: 10.1177/003754970107600201.

[18] E. Rashedi, H. Nezamabadi-Pour, and S. Saryazdi, "GSA: a gravitational search algorithm," *Information sciences,* vol. 179, pp. 2232-2248, 2009.

[19] O. K. Erol and I. Eksin, "A new optimisation method: big bang–big crunch," *Advances in Engineering Software,* vol. 37, pp. 106-111, 2006.

[20] A. Hatamlou, "Black hole: A new heuristic optimisation approach for data clustering," *Information sciences,* 2012.

[21] O. K. Erol and I. Eksin, "A new optimisation method: big bang–big crunch," *Advances in Engineering Software,* vol. 37, pp. 106-111, 2006.

[22] B. Hajek, "A tutorial survey of theory and applications of simulated annealing," 1985 24th IEEE Conference on Decision and Control. IEEE, Dec. 1985. doi: 10.1109/cdc.1985.268599.

[23] X.-S. Yang, "Firefly algorithm, stochastic test functions and design optimisation," *International Journal of Bio-Inspired Computation,* vol. 2, pp. 78-84, 2010.

[24] "Biomimicry of bacterial foraging for distributed optimisation and control," IEEE Control Systems, vol. 22, no. 3. Institute of Electrical and Electronics Engineers (IEEE), pp. 52–67, Jun. 2002. doi: 10.1109/mcs.2002.1004010.

[25] R. Storn and K. Price, "Differential evolution–a simple and efficient heuristic for global optimisation over continuous spaces," *Journal of global optimisation,* vol. 11, pp. 341-359, 1997.

[26] S. Madappady and E. Maben, *Learn ECG in a day*. New Delhi, India: Jaypee Brothers Medical Publishers, 2013.

[27] R. Schmidt and G. Thews, *Human Physiology*. Berlin, Heidelberg: Springer Berlin Heidelberg, 1989.

[28] K. Comerford, *Anatomy & physiology made incredibly easy!*. Philadelphia: Wolters Kluwer Health/Lippincott Williams & Wilkins, 2009.

[29] M. Thaler, *The only EKG book you'll ever need*. 2015.

[30] J. Postlethwait and J. Hopson, *Modern biology*. Orlando: Holt, Rinehart and Winston, 2006.

[31] B. Zaret, M. Moser and L. Cohen, *Heart book*. New York: Hearst Book, 1992.

[32] M. McKinley and V. O'Loughlin, *Human anatomy*. New York: McGraw-Hill, 2012.

[33] N. Singh and S. Singh, "A Modified Mean Gray Wolf Optimization Approach for Benchmark and Biomedical Problems", *Evolutionary Bioinformatics*, vol. 13, p. 117693431772941, 2017. Available: 10.1177/1176934317729413

[34] J. M. Abdullah and T. Ahmed, "Fitness Dependent Optimizer: Inspired by the Bee Swarming Reproductive Process," IEEE Access, vol. 7. Institute of Electrical and Electronics Engineers (IEEE), pp. 43473–43486, 2019. doi: 10.1109/access.2019.2907012.

[35] E. Cortés-Toro, B. Crawford, J. Gómez-Pulido, R. Soto, and J. Lanza-Gutiérrez, "A New Metaheuristic Inspired by the Vapour-Liquid Equilibrium for Continuous Optimisation," Applied Sciences, vol. 8, no. 11. MDPI AG, p. 2080, Oct. 28, 2018. doi: 10.3390/app8112080.

[36] Brest, J., Maučec, M. S., & Bošković, B. (2019, June). The 100-digit challenge: Algorithm jDE100. In *2019 IEEE congress on evolutionary computation (CEC)* (pp. 19-26). IEEE.